\documentclass[10pt,twocolumn,letterpaper]{article}
\usepackage[accsupp]{axessibility}
\usepackage{cvpr}              
\usepackage{float}

\definecolor{cvprblue}{rgb}{0.21,0.49,0.74}
\usepackage[pagebackref,breaklinks,colorlinks,allcolors=cvprblue]{hyperref}

\def\confName{CVPR}
\def\confYear{2026}

\title{Differentiable Stroke Planning with Dual Parameterization for Efficient and High-Fidelity Painting Creation}

\author{Jinfan Liu\textsuperscript{1}, 
~Wuze Zhang\textsuperscript{1}, 
~Zhangli Hu\textsuperscript{1},
~Zhehan Zhao\textsuperscript{1},
~Ye Chen\textsuperscript{1},
~Bingbing Ni\textsuperscript{1}\thanks{Corresponding author.}\\
\textsuperscript{1}Shanghai Jiao Tong University
}

\begin{document}
\maketitle
\begin{abstract}
In stroke-based rendering, search methods often get trapped in local minima due to discrete stroke placement, while differentiable optimizers lack structural awareness and produce unstructured layouts. To bridge this gap, we propose a dual representation that couples discrete polylines with continuous Bézier control points via a bidirectional mapping mechanism. This enables collaborative optimization: local gradients refine global stroke structures, while content-aware stroke proposals help escape poor local optima. Our representation further supports Gaussian-splatting-inspired initialization, enabling highly parallel stroke optimization across the image. Experiments show that our approach reduces the number of strokes by 30–50\%, achieves more structurally coherent layouts, and improves reconstruction quality, while cutting optimization time by 30–40\% compared to existing differentiable vectorization methods.

\end{abstract}    
\vspace{-2mm}
\section{Introduction}
\label{sec:intro}

The pursuit of non-photorealistic rendering, particularly in applications like digital oil painting and interactive sketching, has long been driven by the goal of automatically translating images into expressive, stroke-based artworks. This process, known as stroke-based rendering or image vectorization, fundamentally relies on how strokes are represented and optimized. Existing approaches predominantly fall into two categories, each with significant drawbacks. On one hand, search-based methods \cite{pivie,painterly,gooch,ippr,turk1996,im2oil,spp,sapa} trace strokes by following local pixel gradients, often using discrete, rectangular primitives. This leads to structural fragmentation and sensitivity to image noise. Consequently, these methods typically require a vast number of strokes, often in the tens of thousands for a single image to achieve reasonable fidelity, resulting in slow optimization and mediocre reconstruction quality (e.g., PSNR often below 27 dB for complex textures). On the other hand, differentiable optimization methods \cite{DifferentiableSketching,diffvg,snp} leverage gradient descent for efficient color fitting. While more stable and faster per stroke, they lack explicit structural priors. The optimization is dominated by pixel-level loss, frequently producing visually disorganized and unstructured stroke layouts that fail to capture the coherent flow of image features, thereby tending to be trapped in local minima and limiting their artistic quality and editability.

\begin{figure}[t]
\centering
\includegraphics[width=1.0\linewidth]{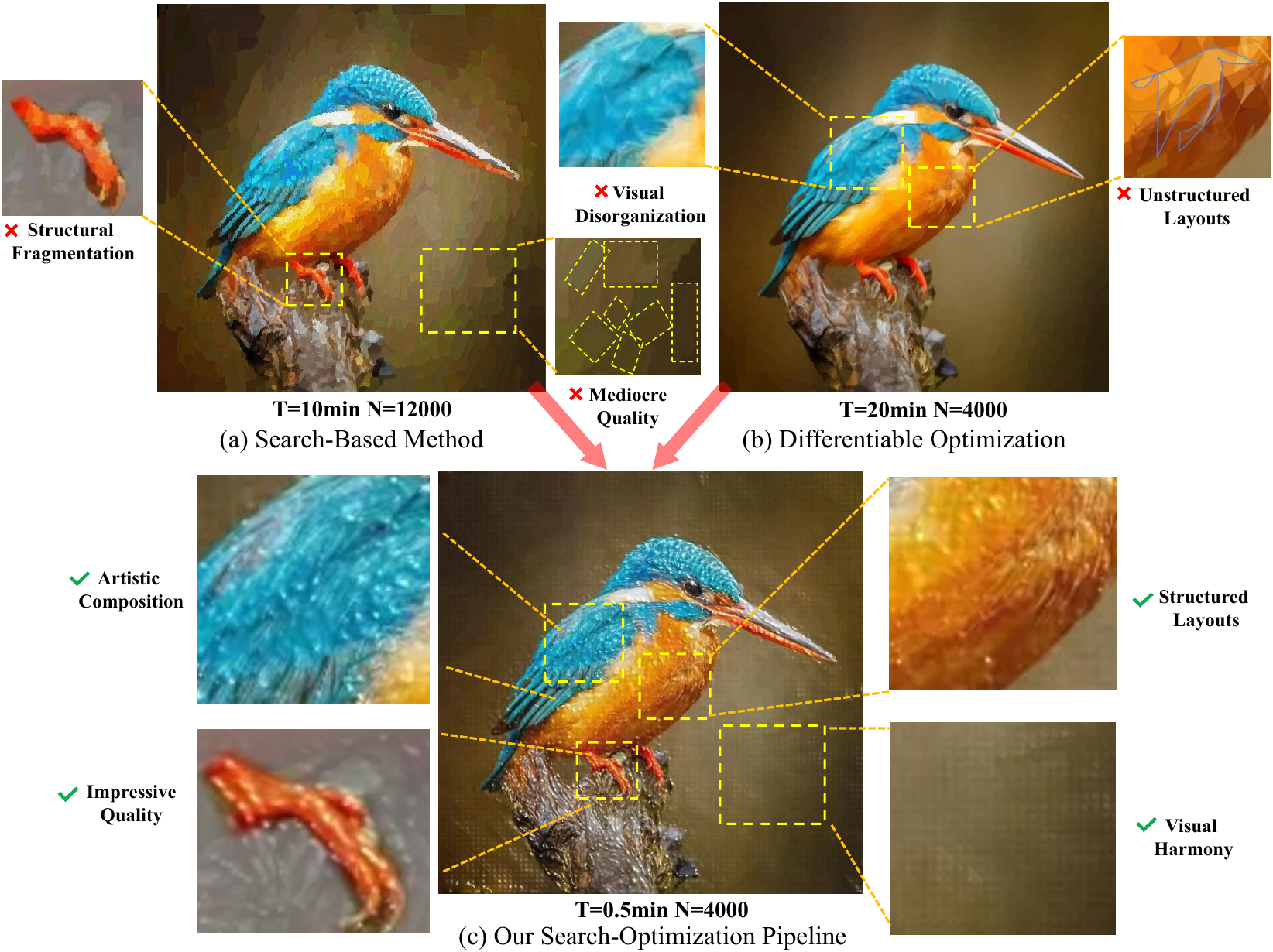}
\caption{
  \textbf{Comparison with existing stroke-based rendering approaches.}
  (a)~Search-based methods suffer from structural fragmentation and require numerous strokes, leading to slow and time-consuming optimization.
  (b)~Differentiable optimization methods converge faster but often produce visually disorganized and unstructured layouts due to the lack of explicit structural priors.
  (c)~Our method combines discrete structure-aware search with differentiable refinement, producing coherent, visually harmonious, and high-fidelity paintings with significantly fewer strokes.
}
\label{fig:motivation}
\vspace{-0.4cm}
\end{figure}

To overcome these inherent limitations, we introduce a novel hybrid optimization framework that seamlessly integrates discrete structural exploration with continuous gradient-based refinement. At its core is a \emph{dual stroke representation}: a discrete polyline defined by a sequence of anchor points, and a continuous counterpart represented by smooth Bézier control points. These two representations are bidirectionally mapped through a differentiable spline conversion layer, allowing loss gradients to flow from the continuous domain back to influence the discrete structure. This coupling enables our two-stage optimization cycle. First, a residual-guided parallel search proactively proposes short, discrete stroke segments in regions with high reconstruction error, effectively capturing major image structures and escaping poor local minima. Second, a differentiable polyline splatting renderer with an anisotropic kernel takes these proposals and refines them using gradient descent, optimizing their color, shape, and transparency for pixel-accurate fitting. A key innovation is our learnable transparency model, which allows strokes to blend smoothly, significantly reducing the number required for high-quality reconstruction. Furthermore, our representation naturally supports a Gaussian-splatting-inspired initialization strategy, where potential stroke seeds are distributed based on image feature density, enabling massive parallel optimization of thousands of strokes from the very beginning. The principal advantages of our system are threefold: 1) it produces structurally coherent and artistically plausible layouts; 2) the resulting
vectorized strokes are highly amenable to editing; and 3) it achieves a significant boost in optimization efficiency and final visual fidelity.

Extensive evaluations across multiple benchmarks show that our method significantly outperforms prior work \cite{im2oil,p-trans,snp,snpsd,p-rl} on a wide range of metrics. With 30–50\% fewer strokes, our approach achieves higher reconstruction fidelity, highlighting its efficient use of primitive elements; on complex texture datasets, our structure-aware optimization yields a 4–5 dB PSNR gain. Importantly, this improvement comes with substantially reduced runtime—thanks to strategic initialization and a GPU-parallel splatting design, the total optimization time decreases by 30–40\%, often converging to high-quality results within minutes. Qualitative comparisons and a user study further confirm that our method produces stroke layouts that are not only more accurate but also visually more coherent and better aligned with salient features such as edges and ridges. Taken together, the combination of efficiency, compact representation, and superior artistic quality marks a meaningful step toward practical, high-quality stroke-based rendering.
\section{Related Work}
\label{sec:related_work}
\subsection{Stroke Planning}
Stroke planning focuses on organizing brush strokes to effectively represent image structures and styles. Early search-based methods \cite{turk1996,pivie,painterly,gooch} trace strokes by following local gradients. Subsequent search-based techniques introduced image segmentation and adaptive sampling to improve reconstruction quality \cite{ippr,spp,sapa,im2oil}. However, these methods often produce structurally fragmented results and require a large number of strokes with slow optimization. Additionally, they remain prone to local minima due to discrete stroke placement. Recently, neural networks like RNNs \cite{Sketch-RNN}, Transformers \cite{p-trans,mambapainter,attentionpainter}, and RL-based approaches \cite{p-rl,continuous-rl,lpaintb,intelli,snpsd} have been adopted for stroke sequence generation, but they face challenges in both generalization and controllability. Meanwhile, differentiable rendering methods \cite{DifferentiableSketching,diffvg,snp,chen2025easy,chen2024towards,chen2023editable} employ gradient-based optimization for efficient color fitting, enabling faster convergence per stroke. However, they often produce visually disorganized layouts due to a lack of structural awareness. Our approach addresses these limitations through a dual representation that combines discrete polylines with continuous Bézier control points.
 
\subsection{Stroke Representation}
Stroke representation is fundamental to stylized brushstroke rendering, as it impacts the visual richness and style fidelity of the output. Early methods such as \cite{pivie,vectorized} model strokes using rectangular segments, which are simple to render but often fail to capture smooth contours, leading to fragmented results and an excessive number of short strokes. To better approximate natural brush trajectories, curved representations have been widely adopted, including B-splines \cite{painterly} and Bézier curves \cite{spp,gpu,offsets,diffvg}. While these support continuous paths and richer shapes, they often suffer from slow rendering and local redundancies, and still lack higher-level structural organization. Other studies enrich stroke appearance using texture templates collected from real artworks \cite{ippr,im2oil}, though such methods frequently exhibit texture discontinuities and limited adaptability. Physics-based modeling techniques simulate pigment-canvas interactions to enhance realism, such as viscous fluid deposition \cite{interactive-hb,impasto} and particle-based pigment dynamics \cite{wetbrush}. These achieve compelling material effects, but rely on complex simulation and are difficult to optimize at the stroke level. 
\begin{figure*}[htbp]
    \centering
    \includegraphics[width=0.98\textwidth]{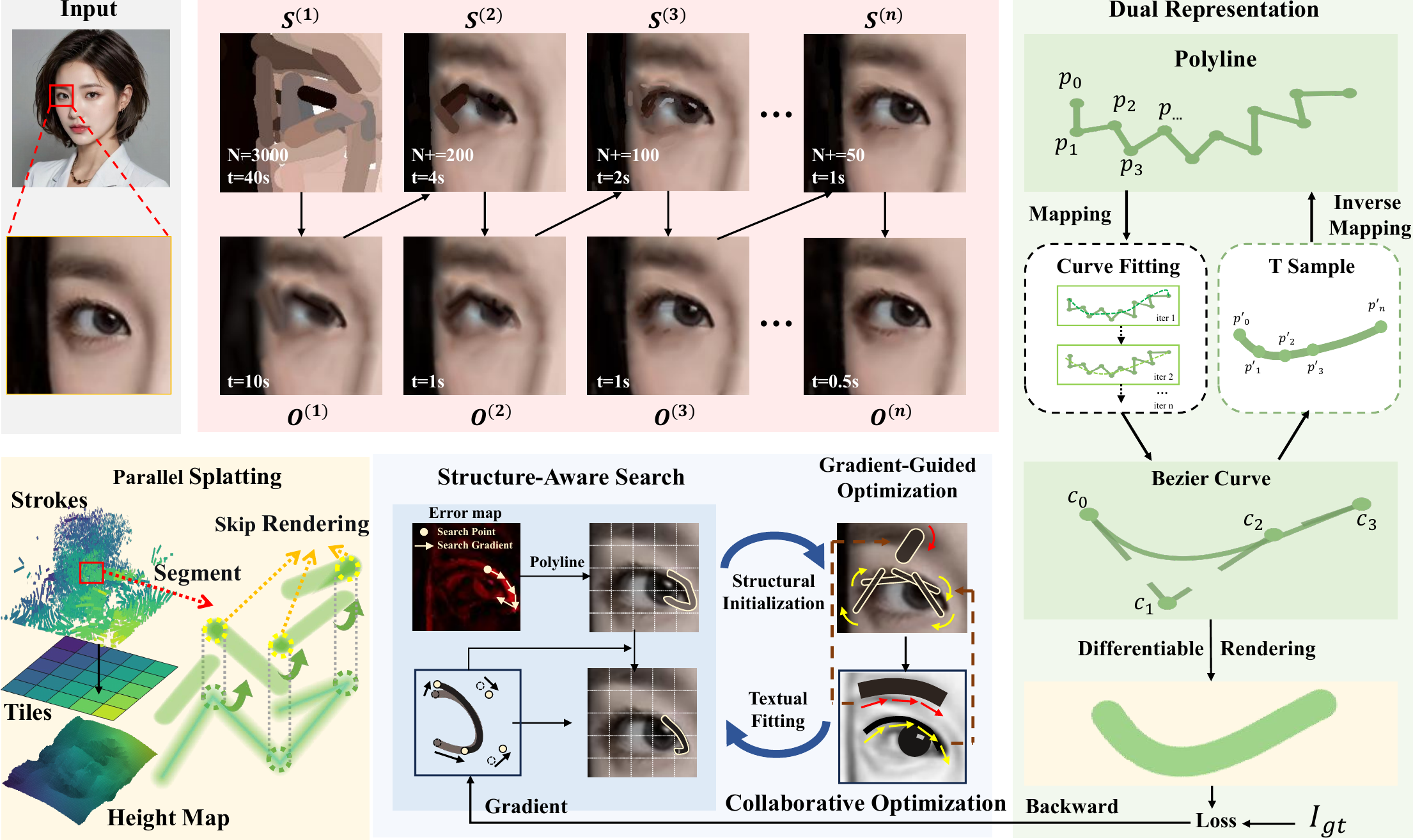}
    \caption{
      \textbf{Framework.}
      Given a target image $I_{\text{gt}}$, our method iteratively reconstructs it through a dual-stage \emph{search-and-optimize} pipeline. At iteration $i$, the structure-aware search step $S(i)$ samples seeds from high-residual regions using NMS and traces geometry-aligned polylines along image gradients. These polylines are then mapped, via a bidirectional differentiable dual representation, to Bézier curves and refined in the optimization step $O(i)$ using gradients from our differentiable polyline splatting renderer. All strokes are splatted in parallel to form the canvas, and the process rapidly converges, with most structure recovered in the first iteration.
      }
    \label{fig:framework}
\end{figure*}

\section{Methodology}
Our method is motivated by a key observation: the complementary strengths of discrete search and continuous optimization can be harnessed through a \textbf{dual stroke representation} that enables them to work in concert. This synergy enables efficient exploration of large-scale structures while achieving precise, pixel-level fidelity.
Given a target image \(I_{\text{gt}} \in \mathbb{R}^{H \times W \times 3}\), our goal is to find a set of strokes \(\mathcal{S} = \{s_i\}\) whose rendering \(I\) minimizes the reconstruction loss \(\mathcal{L}(I, I_{\text{gt}})\). Each stroke \(s_i\) is primarily defined by its spatial trajectory and appearance.

The core of our framework is an iterative \textbf{search-and-optimize} pipeline, built upon our dual stroke representation. As illustrated in Fig. ~\ref{fig:framework}, each iteration consists of two synergistic stages: 1) \textbf{Structure-Aware Search}: Proposes new stroke structures in under-reconstructed regions; 2) \textbf{Gradient-Guided Optimization}: Refines the parameters of all existing strokes. These steps are followed by a \textbf{Differentiable Polyline Renderer} that efficiently composites strokes onto the canvas, along with coherent stroke height regression and lighting. The pipeline repeats until convergence, progressively building a coherent and compact set of strokes that accurately approximates the target image.

\subsection{Dual Stroke Representation}
We introduce a dual representation to bridge the gap between discrete search and continuous gradient optimization, enabling them to collaboratively refine the same stroke to pursue both structural and fine-grained stroke fitting harmony.
To this end, a stroke \(s_i\) is simultaneously represented in both forms:
1) \textbf{Discrete Polyline}: A sequence of ordered vertices, \(\mathbf{P}_i = [\mathbf{p}_{i,0}, \mathbf{p}_{i,1}, \ldots, \mathbf{p}_{i,M_i-1}]\), which is intuitive for structural proposal and manipulation.
2) \textbf{Continuous Bézier Curve}: A smooth parametric curve defined by a set of control points \(\mathbf{C}_i = [\mathbf{c}_{i,0}, \mathbf{c}_{i,1}, \ldots, \mathbf{c}_{i,K-1}]\), which is ideal for differentiable refinement.

These two representations are bidirectionally mapped through a shared parameterization, forming the backbone of our cooperative optimization framework.

\textbf{Mapping: Polyline to Bézier (Differentiable Fitting)}
To convert the discrete polyline \(\mathbf{P}_i\) into a continuous Bézier curve, we first assign a normalized parameter \(t_{i,m} \in [0,1]\) to each vertex \(\mathbf{p}_{i,m}\) using chord-length parameterization along the polyline. The goal is to find the Bézier control points \(\mathbf{C}_i\) that best approximate the polyline vertices. The position of a point on the Bézier curve at parameter \(t\) is given by:
\begin{equation}\small
\mathbf{B}(t; \mathbf{C}_i) = \sum_{k=0}^{K-1} \mathbf{c}_{i,k} \cdot B_{k, K-1}(t),
\end{equation}
where \(B_{k, K-1}(t)\) is the Bernstein basis polynomial of degree \(K-1\). 
We solve for the optimal control points \(\mathbf{C}_i^*\) by minimizing the least-squares error between the original polyline and the fitted curve:
\begin{equation}\small
\mathbf{C}_i^* = \underset{\mathbf{C}_i}{\arg\min} \sum_{m=0}^{M_i-1} \left\| \mathbf{p}_{i,m} - \mathbf{B}(t_{i,m}; \mathbf{C}_i) \right\|^2_2.
\end{equation}
This convex problem has a closed-form solution via the pseudoinverse, allowing the fitting process to be trivially differentiated, thus enabling gradient flow from the continuous curve back to the polyline vertices.

\textbf{Mapping: Bézier to Polyline (Differentiable Sampling)}
The reverse mapping is straightforward and fully differentiable. Given a set of Bézier control points \(\mathbf{C}_i\), we can sample the corresponding discrete polyline \(\mathbf{P}_i\) by evaluating the Bézier curve at a fixed set of parameters \(\{t_{i,m}\}\):
\begin{equation}\small
\mathbf{p}_{i,m} = \mathbf{B}(t_{i,m}; \mathbf{C}_i)
\end{equation}
This sampling operation is a linear combination of the control points and thus differentiable with respect to \(\mathbf{C}_i\).

This bidirectional, differentiable bridge allows gradient signals from the continuous optimization to refine the discrete structure proposed by the search module, ensuring that both global layout and local details are optimized.

\subsection{Structure-Aware Stroke Search}
\label{sec:Search}
This module is responsible for proactively discovering strokes that capture prominent image structures. The core of our search is a \textbf{gradient-driven tracing} mechanism, which directly leverages the continuous gradient field to guide the discrete growth of polylines, ensuring that the proposed structures are both geometrically salient and optimized for loss reduction from their inception.

The search begins with the reconstruction residual \(R^t\) and its gradient field \(\mathbf{G}^t\) at iteration \(t\):
\begin{equation}\small
R^t(\mathbf{x}) = \|I^{t}(\mathbf{x}) - I_{\text{gt}}(\mathbf{x})\|_2^2, \quad 
\mathbf{G}^t(\mathbf{x}) = \nabla R^t(\mathbf{x}).
\end{equation}
Seed points \(\{\mathbf{x}_j\}\) are sampled from the distribution of \(R^t\) using non-maximum suppression. For each seed, rather than merely initializing a fixed-direction segment, we \textbf{trace a polyline} by iteratively following the path that minimizes the reconstruction error.

Starting from \(\mathbf{p}_0 = \mathbf{x}_j\), we grow the polyline by repeatedly stepping along a gradient-flow direction \(\mathbf{d}_k\):
\begin{equation}\small
\mathbf{p}_{k+1} = \mathbf{p}_k + \eta \cdot \mathbf{d}_k ,
\end{equation}
where \(\eta\) is an adaptive step size. Intuitively, \(\mathbf{d}_k\) performs a \textbf{gradient ascent} on the residual field, pushing the stroke to cover high-error regions.

To prevent the trace from being trapped by local noise, the direction \(\mathbf{d}_k\) at each step is a weighted combination of the local gradient and the previous direction \(\mathbf{d}_{k-1}\), encouraging smoothness:
\begin{equation}\small
\tilde{\mathbf{d}}_k = \frac{\mathbf{G}^t(\mathbf{p}_k)}{\|\mathbf{G}^t(\mathbf{p}_k)\| + \epsilon}, \quad
\mathbf{d}_k = \lambda \cdot \tilde{\mathbf{d}}_k + (1-\lambda) \cdot \mathbf{d}_{k-1}.
\end{equation}

The tracing terminates when the current stroke no longer reduces the reconstruction loss, and the overall search exceeds a threshold of consecutive failures.

This method produces a set of discrete polylines \(S_{\text{search}} = \{\mathbf{P}_i\}\) where each vertex is \emph{explicitly dictated by the gradient flow}. The resulting strokes are therefore not only structurally coherent but also primed for loss reduction. They are then directly passed through our dual representation to become Bézier curves, ready for fine-grained, differentiable adjustment in the next stage. In this way, the gradient field serves as a continuous guide throughout the entire discrete search process, achieving true collaboration between the two optimization paradigms.

\subsection{Gradient-Guided Stroke Optimization}
\label{sec:Optimization}
The stroke set \(S_{\text{search}} = \{\mathbf{P}_i\}\) proposed by the discrete search module provides a structurally coherent initialization in the form of discrete polylines. This module refines the set within the \textbf{continuous Bézier representation} \(\{\mathbf{C}_i\}\) to achieve pixel-level fidelity. The key to our efficient co-design is that the search module has already solved the large-scale structural assignment, allowing the gradient-based optimizer to focus on fine-grained adjustments without the risk of collapsing into globally poor layouts.

All strokes are optimized concurrently. The Bézier control points \(\mathbf{C}_i\) for each stroke are uniformly sampled to produce a dense, differentiable polyline \(\tilde{\mathbf{P}}_i\) for rendering. The entire set is then rendered in a single forward pass using our differentiable splatting kernel (Sec.~\ref{sec:renderer}) to produce the canvas \(I_{\text{render}}\). We minimize the reconstruction objective
\begin{equation}\small
\mathcal{L}=\| I_{\text{render}} - I_{\text{gt}} \|_2
+ \lambda_{\ell}\mathcal{L}_{\text{len}}
+ \lambda_{w}\mathcal{L}_{\text{width}},
\end{equation}
where \(\mathcal{L}_{\text{len}}\) and \(\mathcal{L}_{\text{width}}\) regularize stroke length and width. Gradients \(\nabla_{\mathbf{C}_i, \mathbf{c}_i, \mathbf{o}_i, w_i}\mathcal{L}\) are fully backpropagated through the renderer, enabling end-to-end refinement of geometry, color, opacity, and width.

We employ an adaptive optimization strategy similar to 3DGS \cite{3dgs} for stable convergence. Since our strokes are typically long and structurally important, directly splitting or pruning them would damage global contour coherence.
 We therefore only reinitialize strokes with near-zero opacity and minimal reconstruction loss, letting the structure-aware search module assign their new parameters. This preserves structural integrity while ensuring stable and efficient optimization.

This continuous refinement directly benefits from the structural prior provided by the search stage. Since the initial strokes are already well-placed along salient edges and ridges, the gradient descent efficiently fine-tunes their geometry and appearance rather than attempting costly global reconfiguration. This division of labor—\textbf{coarse exploration via search, fine exploitation via gradients}—is the cornerstone of our efficiency.

Conversely, the gradient optimization legitimizes and enhances the proposals from the discrete search. A search-proposed stroke might be geometrically approximate; the gradient refinement \textbf{corrects} its precise placement, width, and color, transforming a structurally correct but coarse proposal into a visually accurate and seamless element of the final painting. This synergistic cycle iterates until convergence, progressively building a compact and coherent stroke set \(S^{*}\).

\subsection{Differentiable Polyline Renderer}
\label{sec:renderer}
To enable end-to-end optimization while maintaining high rendering efficiency, we design a differentiable renderer that approximates each stroke segment as a soft, anisotropic splat. This offers a favorable trade-off between the computational cost of exact Bézier rasterization and the need for full differentiability.

Each stroke, after being sampled from its Bézier representation into a polyline \(\tilde{\mathbf{P}}_i\), is decomposed into consecutive segments. The influence of a segment \(e_{i,j}\) on a pixel \(\mathbf{x}\) is modeled by an anisotropic kernel based on the shortest distance \(d_{i,j}(\mathbf{x})\):
\begin{equation}\small
k_{i,j}(\mathbf{x}) = \frac{\sigma\left(\frac{w_i / 2 - d_{i,j}(\mathbf{x})}{\tau}\right) - \sigma\left(\frac{-w_i / 2}{\tau}\right)}{1 - 2 \cdot \sigma\left(\frac{-w_i / 2}{\tau}\right)},
\end{equation}
where \(w_i\) is the stroke width, \(\tau\) a softness parameter, and \(\sigma\) the logistic function. The resulting opacity is \(\alpha_{i,j}(\mathbf{x}) = \mathbf{o}_{i} \cdot \max(0, k_{i,j}(\mathbf{x}))\).

All segments are composited onto the canvas in a single, GPU-parallel pass using front-to-back alpha blending. Let \(\pi(\mathbf{x})\) denote the sequence of stroke segments overlapping pixel \(\mathbf{x}\), sorted from front to back:
\begin{equation}\small
C(\mathbf{x}) = \sum_{(i,j) \in \pi(\mathbf{x})} \alpha_{i,j}(\mathbf{x}) \, \mathbf{c}_{i} \prod_{(p,q) < (i,j)} (1 - \alpha_{p,q}(\mathbf{x})).
\label{accu_color}
\end{equation}

A key to our efficiency is that this splatting paradigm, inspired by recent advances in Gaussian splatting, enables highly parallelized rasterization of all stroke segments. 
This directly enables the \textbf{Gaussian-splatting-inspired initialization} in Sec.~\ref{sec:Optimization}, where a large population of stroke primitives can be placed and optimized simultaneously from the first iteration, dramatically accelerating convergence.

The entire rendering process is fully differentiable, allowing gradients to flow from the total loss \(\mathcal{L}\) back to the stroke parameters \(\mathbf{C}_i\), \(\mathbf{c}_i\), \(\mathbf{o}_i\), and \(w_i\). This efficient, parallel, and differentiable renderer is the backbone that makes our iterative search-and-optimize pipeline practical.

To further enhance the physical realism of the painting, we regress a height value \(h_i\) for each stroke, enabling advanced shading effects via relighting. The initial height is estimated from the target image using a pre-trained monocular depth estimation model (e.g., Depth Anything~\cite{yang2024depth}), which provides a dense geometry prior. This initial prediction is then refined with a continuity constraint, encouraging smooth height transitions between spatially adjacent strokes. The final height field allows us to render the painting under novel lighting conditions, simulating the tangible surface relief characteristic of impasto oil paintings.

\begin{figure*}[htbp]
    \centering
    \includegraphics[width=\linewidth]{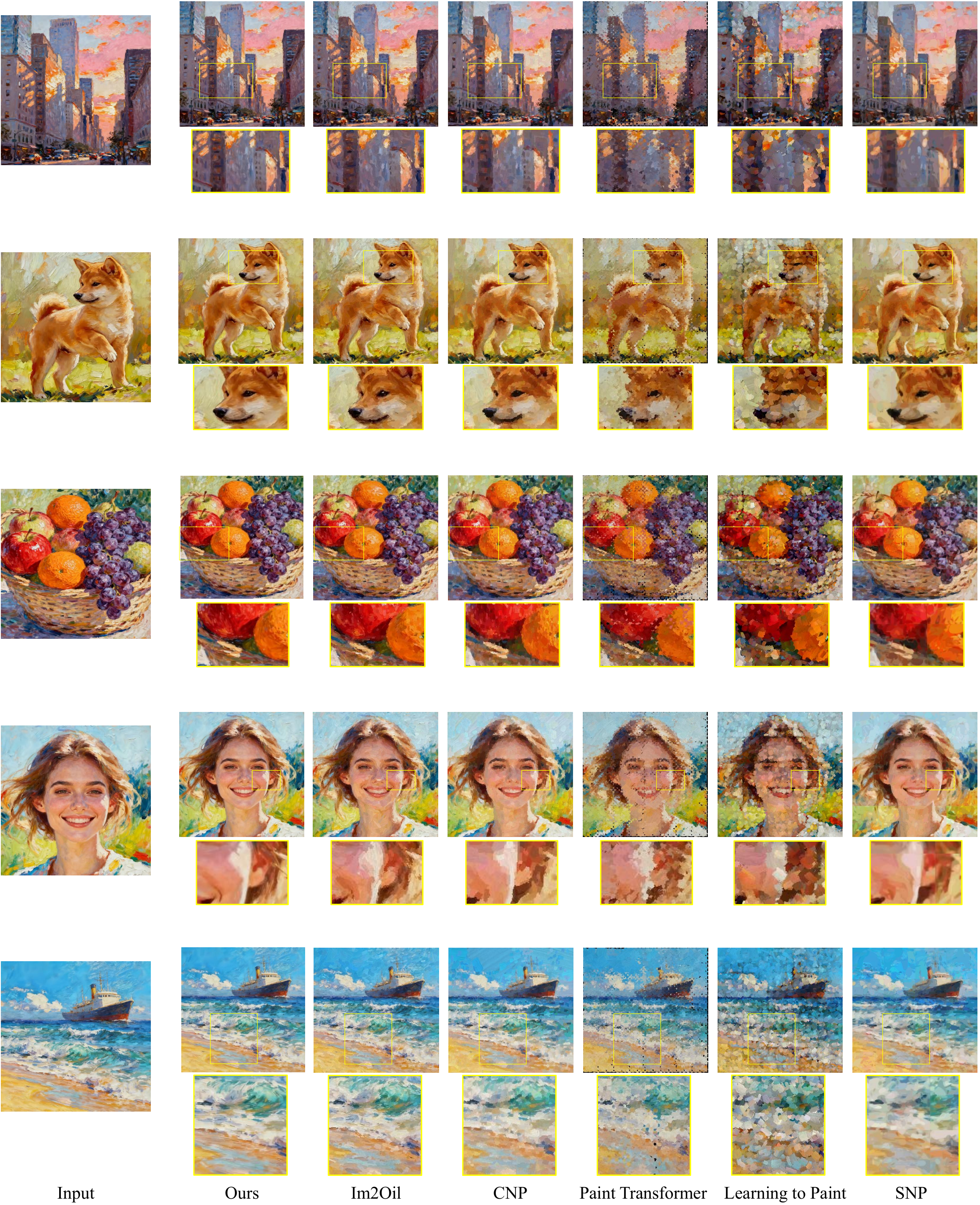}
    \caption{
        \textbf{Qualitative comparison with state-of-the-art methods.}
        Our method produces clearer contours, cleaner textures, and more coherent stroke flows across both natural and synthetic domains. 
        In contrast, baseline approaches often generate noisy strokes or lose structural consistency, particularly under limited stroke budgets.
    }
    \label{fig:qualitative}
\end{figure*}
\section{Experiments}

\subsection{Experimental Setups}
\label{exp:setup}   
\noindent\textbf{Implementation details.}
All experiments are implemented in PyTorch and run on a single NVIDIA RTX 4090 GPU. 
Since our pipeline already achieves strong convergence in the first iteration, we use a default of three search–optimization iterations. 
During the \textbf{search phase}, we extract seeds from the top 12\% residual pixels using a $7\!\times\!7$ NMS window. We trace strokes using an adaptive step, starting at 1.2 pixels with a smoothing factor of $\lambda=0.8$. Strokes grow up to 20 vertices dynamically and are accepted only if they reduce the reconstruction loss by at least $0.01$. The local search halts after 20 consecutive rejections.

During the \textbf{optimization phase}, discovered polylines are converted into piecewise cubic Bézier curves and sampled at 10 discrete points for differentiable rendering. We jointly optimize all stroke parameters using Adam for up to 4000 iterations per stage or until convergence. The base learning rate is 0.01, with color and width updates scaled by 0.1 and 0.01, respectively.

\noindent\textbf{Dataset \& Evaluation.}
We evaluate on two complementary datasets. 
(1) The DIV2K validation set~\cite{div2k}, a high-resolution and diverse natural-image benchmark resized to 1200$\times$1200 for consistent stroke resolution. 
(2) The Im2Oil gallery dataset~\cite{im2oil}, featuring classical oil paintings at $600 \times 800$ with characteristic textures and styles. We report PSNR, SSIM, LPIPS, stroke count, per-iteration runtime, and a perceptual user study.

\subsection{Comparative Analysis with SOTA}
\label{exp:expA}
We compare our approach with five state-of-the-art stroke-based rendering methods, including Im2Oil~\cite{im2oil}, Compositional Neural Painting (CNP)~\cite{snpsd}, Paint Transformer~\cite{p-trans}, Learning to Paint~\cite{p-rl}, and Stylized Neural Painting (SNP)~\cite{snp}. We employ 16K strokes for the DIV2K dataset and 8K strokes for the Im2Oil dataset based on image resolution, to align parameters across all baselines. 

Qualitative comparisons in Fig.~\ref{fig:qualitative} show that our method consistently produces clearer structural boundaries, cleaner textures, and more coherent stroke flows on artistic images. By leveraging our Dual Stroke Representation and joint search-optimization pipeline, we effectively preserve thin structures and smooth regions, whereas baselines struggle with noisy, overlapping strokes. 

Quantitatively, Table~\ref{tab:main} shows our superior reconstruction accuracy across both datasets. Notably, the substantial improvements in SSIM highlight our method's capability in preserving global structures and semantic coherence. Beyond fidelity, our method exhibits remarkable computational efficiency. By synergizing the dual stroke representation with GPU-parallel differentiable splatting, our pipeline substantially accelerates the optimization process, running significantly faster than existing SBR baselines.

To assess perceptual quality beyond pixel-level metrics, we conducted a user study with 100 participants, including 60 from a university campus and 40 from an online platform; about 30\% had formal training in visual arts or design.
In each trial, participants were shown a set of source images together with six corresponding oil paintings generated by our method and five baselines, displayed in randomized order. They were asked to select the best painting and then marked the regions they found most compelling by drawing bounding boxes, encouraging fine-grained judgments. Participants further rated each painting across four perceptual criteria commonly used in non-photorealistic rendering: \emph{Structural Fidelity}, \emph{Textural Realism}, \emph{Color Accuracy}, and \emph{Overall Preference}, each on a 1–5 Likert scale. Table~\ref{tab:userstudy} shows that our method ranks highest across all criteria, confirming its perceptual superiority.

\begin{table}[t]
\centering
\resizebox{\linewidth}{!}{
\setlength{\tabcolsep}{5pt}
\begin{tabular}{lcccc}
\toprule
\textbf{Method} & PSNR↑ & SSIM↑ & LPIPS↓ & Time (s)↓ \\
\midrule
\multicolumn{5}{l}{\textit{DIV2K Validation}} \\
\midrule
Im2Oil~\cite{im2oil}               & 27.59 & 0.72 & 0.211 & 727.8 \\
CNP~\cite{snpsd}                 & 27.91 & 0.64 & 0.296 & 125.5 \\
Paint Transformer~\cite{p-trans}   & 23.83 & 0.63 & 0.378 & 414.1 \\
Learning to Paint~\cite{p-rl}        & 27.19 & 0.70 & 0.330 & 167.9 \\
SNP~\cite{snp}                     & 20.63 & 0.42 & 0.405 & 5837.2 \\
\textbf{Ours}                      & \textbf{32.16} & \textbf{0.93} & \textbf{0.076} & \textbf{87.6} \\
\midrule
\multicolumn{5}{l}{\textit{Gallery Dataset}} \\
\midrule
Im2Oil~\cite{im2oil}               & 28.54 & 0.71 & 0.204 & 392.8 \\
CNP~\cite{snpsd}                 & 28.46 & 0.68 & 0.288 & 61.3 \\
Paint Transformer~\cite{p-trans}   & 24.46 & 0.61 & 0.395 & 197.4 \\
Learning to Paint~\cite{p-rl}        & 28.58 & 0.73 & 0.318 & 80.6 \\
SNP~\cite{snp}                     & 20.87 & 0.49 & 0.433 & 2868.1 \\
\textbf{Ours}                      & \textbf{32.53} & \textbf{0.86} & \textbf{0.192} & \textbf{42.7} \\
\bottomrule
\end{tabular}}
\caption{Quantitative comparison with SOTA SBR methods.}
\label{tab:main}
\vspace{-2mm}
\end{table}

\begin{table}[t]
\centering
\resizebox{\linewidth}{!}{
\setlength{\tabcolsep}{5pt}
\begin{tabular}{lcccc}
\toprule
\textbf{Method} &
\textbf{Structure↑} &
\textbf{Texture↑} &
\textbf{Color↑} &
\textbf{Overall↑} \\
\midrule
Im2Oil~\cite{im2oil}               & 3.42 & 3.28 & 3.55 & 3.38 \\
CNP~\cite{snpsd}                   & 3.65 & 3.12 & 3.08 & 3.31 \\
Paint Transformer~\cite{p-trans}   & 2.98 & 3.05 & 2.87 & 2.97 \\
Learning to Paint~\cite{p-rl}      & 3.51 & 3.42 & 3.47 & 3.47 \\
SNP~\cite{snp}                     & 2.45 & 2.38 & 2.31 & 2.41 \\
\midrule
\textbf{Ours}                      & \textbf{4.32} & \textbf{4.28} & \textbf{4.41} & \textbf{4.35} \\
\bottomrule
\end{tabular}}
\caption{User study results (1–5 Likert ratings; higher is better). Our method achieves the best rank across all perceptual criteria.}
\label{tab:userstudy}
\vspace{-2mm}
\end{table}

\subsection{Tunable Aesthetics and Physical Realism}
\label{exp:texture}
\begin{figure}[t]
    \centering
    \includegraphics[width=\linewidth]{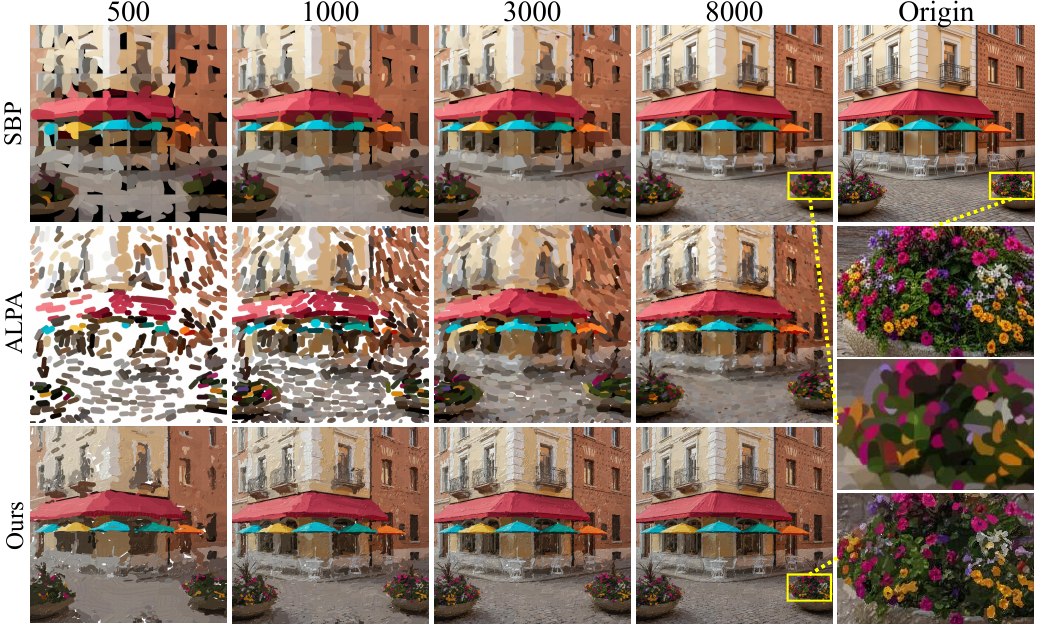}
    \vspace{-4mm} 
    \caption{
        Progressive painting process with height-field relighting.
    }
    \label{fig:evolution}
    \vspace{-4mm}
\end{figure}

Beyond reconstruction fidelity, our pipeline offers highly tunable aesthetics and physical realism. Stroke appearance is flexibly controlled via the hardness parameter $\tau$: higher values ($\tau=0.7$, evaluated in Sec.~\ref{exp:expA}) yield smooth, blended photorealism, whereas lower values ($\tau=0.1$) produce sharp, distinct brush-like strokes that vividly reveal the structural layout. 
Under this hard-edge setting, Fig.~\ref{fig:evolution} compares our progressive rendering against ALPA~\cite{sapa} and SBP~\cite{spp}, two recent semantic SBR baselines. Beyond achieving superior fidelity, structural coherence, and stroke efficiency, our method transcends standard 2D image generation.
By rendering splatting-based height fields with physics-based lighting, we produce a volumetric 3D model with tangible surface relief, maintaining expressive stroke flows and strong structural coherence even in this physically realistic, hard-edge regime.

\subsection{Ablation Study}
\label{exp:ablation}
\subsubsection{Analysis of Iterative Optimization Dynamics}
\label{exp:expB}
Unless otherwise specified, all ablation experiments are conducted on the DIV2K dataset.
The core of our method lies in the \textit{Iterative Optimization Dynamics}, for which we set the number of iterations to $t=3$ in our default configuration. In this section, we conduct ablation experiments to evaluate both the effectiveness and efficiency of the iterative process. As shown in Table~\ref{tab:iter_vis}, we compare the painting reconstruction quality and runtime cost for $t=1$ through $t=4$. When $t=1$, our method already achieves highly competitive reconstruction performance in under 50 seconds. Increasing the number of iterations further improves reconstruction quality with only a marginal increase in optimization time, and we find $t=3$ to offer the best trade-off between performance and efficiency.

\begin{figure}[t]
\centering
\includegraphics[width=\linewidth]{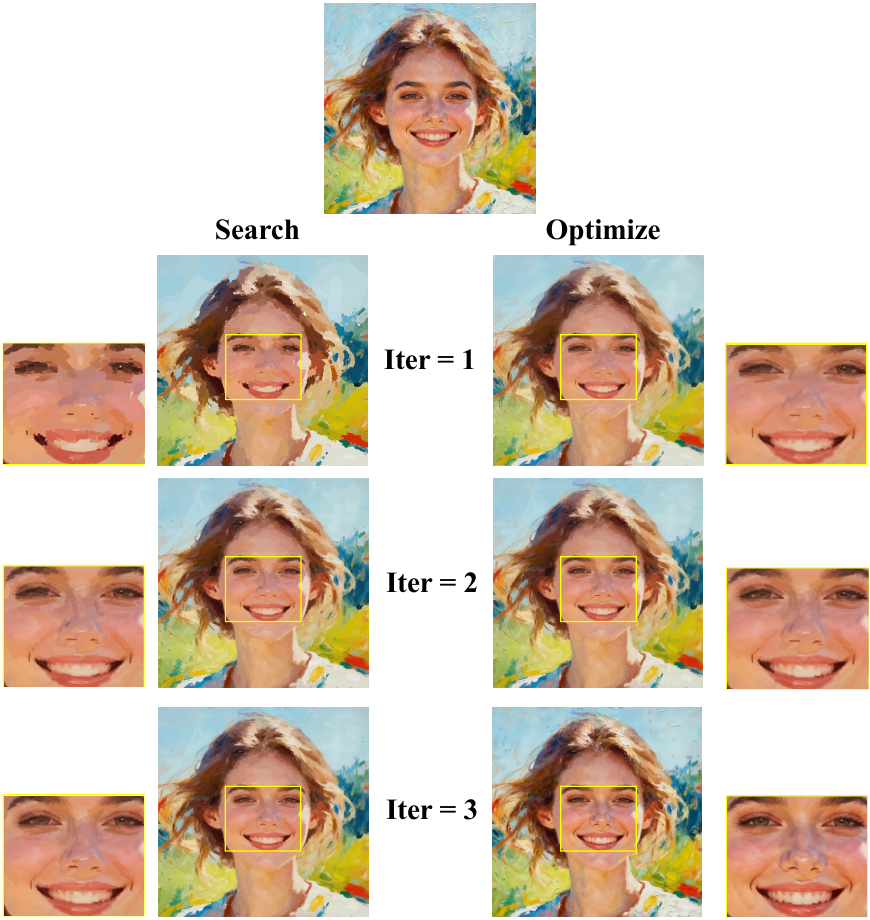}
\caption{
Qualitative visualization of intermediate canvases at iterations $t=\{1,2,3\}$. 
Iteration~1 establishes coarse structural coverage via search, while later iterations progressively refine geometric precision, stroke smoothness, and color consistency.
}
\label{fig:iter_vis}
\vspace{-3mm}
\end{figure}

The qualitative results in Figure.~\ref{fig:iter_vis} further validate these observations. With just a single iteration, our approach already produces stroke-based renderings with strong artistic style and visual realism. As the number of iterations increases, the generated paintings progressively refine geometric precision, stroke smoothness, and color consistency.

\begin{table}[t]
\centering
\resizebox{\linewidth}{!}{
\setlength{\tabcolsep}{5pt}
\begin{tabular}{cccccc}
\toprule
\textbf{Iter.} & \textbf{Stage} & \textbf{PSNR↑} & \textbf{SSIM↑} & \textbf{LPIPS↓} & \textbf{Time (s)} \\
\midrule

\multirow{2}{*}{$t=1$}
    & Search      & 20.3 & 0.48 & 0.323 & 35.6 \\ 
    & Optimize    & 28.4 & 0.82 & 0.214 & 47.2 \\ 
\midrule

\multirow{2}{*}{$t=2$}
    & Search      & 29.6 & 0.83 & 0.215 & 59.0 \\
    & Optimize    & 30.4 & 0.85 & 0.194 & 65.5 \\
\midrule

\multirow{2}{*}{$t=3$}
    & Search      & 29.9 & 0.84 & 0.197 & 75.6 \\
    & Optimize    & 32.2 & 0.93 & 0.076 & 87.6 \\
\midrule

\multirow{2}{*}{$t=4$}
    & Search      & 31.6 & 0.89 & 0.112 & 89.6 \\
    & Optimize    & 32.4 & 0.93 & 0.073 & 95.2 \\
\bottomrule
\end{tabular}}
\caption{Iteration-by-iteration analysis.}
\label{tab:iter_vis}
\vspace{-1mm}
\end{table}

\subsubsection{Effectiveness of the Dual Stroke Representation}
\label{exp:expC}

We evaluate the impact of our dual stroke representation by comparing four variants:  
1) \textit{Search-only}, where strokes are produced by the structure-aware search but not refined;  
2) \textit{Polyline-only optimization}, which optimizes polyline vertices without conversion to Bézier curves;  
3) \textit{Bézier-only optimization}, where Bézier refinement is performed without discrete search initialization;
4) \textit{Full model}, our joint search–optimization pipeline using the dual representation.

Due to the prohibitive rendering cost of pure Bézier optimization, we limit the Bézier-only variant to 2,000 strokes, while other variants use 16,000 strokes. As shown in Table~\ref{tab:representation}, search-only reconstruction captures coarse structure but lacks fine detail; polyline-only optimization suffers from unstable gradients; Bézier-only optimization improves smoothness but often misses global structure due to the absence of discrete guidance. In contrast, our full model consistently achieves the highest fidelity and the lowest perceptual error, demonstrating that discrete search provides strong structural priors while Bézier curves offer a well-behaved space for continuous refinement. These results confirm that the dual representation is essential for both accuracy and stability.

\begin{table}[t]
\centering
\resizebox{\linewidth}{!}{
\setlength{\tabcolsep}{5pt}
\begin{tabular}{lcccc}
\toprule
\textbf{Variant} & PSNR↑ & SSIM↑ & LPIPS↓ & Time (s)↓ \\
\midrule
Search-only                   & 27.8 & 0.76 & 0.227 & 57.2 \\
Polyline-only optimization    & 31.2 & 0.89 & 0.126 & \textbf{31.6} \\
Bézier-only optimization      & 30.4 & 0.86 & 0.138 & 3978.4 \\
\textbf{Full model (ours)}    & \textbf{32.16} & \textbf{0.93} & \textbf{0.076} & 87.6 \\
\bottomrule
\end{tabular}}
\vspace{-1mm}
\caption{Ablation on the dual stroke representation.}
\label{tab:representation}
\vspace{-2mm}
\end{table}

\subsubsection{Ablation on Search and Optimization Components}
As detailed in Table~\ref{tab:ablation}, we ablate key components of our pipeline. For the search module, removing residual-guided seeding fails to target salient regions, while disabling gradient-driven tracing yields fragmented stroke geometry. Both are essential for structurally meaningful proposals. 
For the optimization module, disabling opacity reinitialization retains visually negligible strokes, hindering convergence, while removing loss reinitialization fails to correct misaligned geometry. Eliminating both yields the worst performance due to persistently collapsed strokes. Conversely, our adaptive approach selectively resets underperforming strokes, yielding stable refinement and yields optimal fidelity. 
Ultimately, the search stage provides strong structural priors, and adaptive reinitialization ensures stable gradient-based refinement.

\begin{table}[t]
\centering
\small
\resizebox{\linewidth}{!}{
\begin{tabular}{lccc}
\toprule
\textbf{Variant} & PSNR $\uparrow$ & SSIM $\uparrow$ & LPIPS $\downarrow$ \\
\midrule
\multicolumn{4}{l}{\textit{Search components}} \\
w/o residual seeding    & 26.6 & 0.68 & 0.263 \\
w/o gradient tracing    & 26.3 & 0.71 & 0.249 \\
\textbf{Full search}    & \textbf{27.8} & \textbf{0.76} & \textbf{0.227} \\
\midrule
\multicolumn{4}{l}{\textit{Optimization components}} \\
w/o opacity reinit      & 30.3 & 0.85 & 0.158 \\
w/o loss reinit         & 29.9 & 0.86 & 0.152 \\
w/o any reinit          & 28.1 & 0.83 & 0.173 \\
\textbf{Full optimization} & \textbf{31.2} & \textbf{0.89} & \textbf{0.126} \\
\bottomrule
\end{tabular}
}
\caption{Ablation study on search and optimization components.}
\label{tab:ablation}
\end{table}
\section{Conclusion}
\label{sec:conclu}


We couple discrete polylines with continuous Bézier curves via a differentiable bidirectional mapping, allowing discrete search to establish global structure while gradients refine local details. Integrated with our splatting-based renderer, this approach enables fast optimization and yields high-fidelity painterly renderings using height-based shading.

\noindent\textbf{Acknowledgement.} 
This work is supported by the Science and Technology Commission of Shanghai Municipality under research grant No. 25ZR1401187.

{
    \small
    \bibliographystyle{ieeenat_fullname}
    \bibliography{main}

@String(CVPR= {IEEE Conf. Comput. Vis. Pattern Recog.})

@String(TOG= {ACM Trans. Graph.})

@String(CVPR  = {CVPR})

@String(TOG   = {ACM TOG})

@inproceedings{pivie,
  title={Processing images and video for an impressionist effect},
  author={Litwinowicz, Peter},
  booktitle={Proceedings of the 24th annual conference on Computer graphics and interactive techniques},
  pages={407--414},
  year={1997}
}

@inproceedings{turk1996,
  title={Image-guided streamline placement},
  author={Turk, Greg and Banks, David},
  booktitle={Proceedings of the 23rd annual conference on Computer graphics and interactive techniques},
  pages={453--460},
  year={1996}
}

@inproceedings{gooch,
  title={Artistic vision: painterly rendering using computer vision techniques},
  author={Gooch, Bruce and Coombe, Greg and Shirley, Peter},
  booktitle={Proceedings of the 2nd international symposium on Non-photorealistic animation and rendering},
  pages={83--ff},
  year={2002}
}

@article{ippr,
  title={From image parsing to painterly rendering.},
  author={Zeng, Kun and Zhao, Mingtian and Xiong, Caiming and Zhu, Song Chun},
  journal={ACM Trans. Graph.},
  volume={29},
  number={1},
  pages={2--1},
  year={2009}
}

@inproceedings{spp,
  title={Segmentation-based parametric painting},
  author={de Guevara, Manuel Ladron and Fisher, Matt and Hertzmann, Aaron},
  booktitle={2024 IEEE International Conference on Multimedia and Expo Workshops (ICMEW)},
  pages={1--6},
  year={2024},
  organization={IEEE}
}

@inproceedings{sapa,
  title={Towards artist-like painting agents with multi-granularity semantic alignment},
  author={Hu, Zhangli and Chen, Ye and Zhao, Zhongyin and Liu, Jinfan and Ke, Bilian and Ni, Bingbing},
  booktitle={Proceedings of the 32nd ACM International Conference on Multimedia},
  pages={10191--10199},
  year={2024}
}

@inproceedings{im2oil,
  title={Im2oil: Stroke-based oil painting rendering with linearly controllable fineness via adaptive sampling},
  author={Tong, Zhengyan and Wang, Xiaohang and Yuan, Shengchao and Chen, Xuanhong and Wang, Junjie and Fang, Xiangzhong},
  booktitle={Proceedings of the 30th ACM international conference on multimedia},
  pages={1035--1046},
  year={2022}
}

@article{Sketch-RNN,
  title={A neural representation of sketch drawings},
  author={Ha, David and Eck, Douglas},
  journal={arXiv preprint arXiv:1704.03477},
  year={2017}
}

@inproceedings{p-trans,
  title={Paint transformer: Feed forward neural painting with stroke prediction},
  author={Liu, Songhua and Lin, Tianwei and He, Dongliang and Li, Fu and Deng, Ruifeng and Li, Xin and Ding, Errui and Wang, Hao},
  booktitle={Proceedings of the IEEE/CVF international conference on computer vision},
  pages={6598--6607},
  year={2021}
}

@incollection{mambapainter,
  title={MambaPainter: Neural stroke-based rendering in a single step},
  author={Sawada, Tomoya and Katsurai, Marie},
  booktitle={SIGGRAPH Asia 2024 Posters},
  pages={1--2},
  year={2024}
}

@article{attentionpainter,
  title={AttentionPainter: an efficient and adaptive stroke predictor for scene painting},
  author={Tang, Yizhe and Wang, Yue and Hu, Teng and Yi, Ran and Tan, Xin and Ma, Lizhuang and Lai, Yu-Kun and Rosin, Paul L},
  journal={IEEE Transactions on Visualization and Computer Graphics},
  year={2025},
  publisher={IEEE}
}

@inproceedings{p-rl,
  title={Learning to paint with model-based deep reinforcement learning},
  author={Huang, Zhewei and Heng, Wen and Zhou, Shuchang},
  booktitle={Proceedings of the IEEE/CVF international conference on computer vision},
  pages={8709--8718},
  year={2019}
}

@article{continuous-rl,
  title={Continuous control with deep reinforcement learning},
  author={Lillicrap, Timothy P and Hunt, Jonathan J and Pritzel, Alexander and Heess, Nicolas and Erez, Tom and Tassa, Yuval and Silver, David and Wierstra, Daan},
  journal={arXiv preprint arXiv:1509.02971},
  year={2015}
}

@article{lpaintb,
  title={Lpaintb: Learning to paint from self-supervision},
  author={Jia, Biao and Brandt, Jonathan and Mech, Radom{\'\i}r and Kim, Byungmoon and Manocha, Dinesh},
  journal={arXiv preprint arXiv:1906.06841},
  year={2019}
}

@inproceedings{intelli,
  title={Intelli-Paint: Towards developing more human-intelligible painting agents},
  author={Singh, Jaskirat and Smith, Cameron and Echevarria, Jose and Zheng, Liang},
  booktitle={European conference on computer vision},
  pages={685--701},
  year={2022},
  organization={Springer}
}

@inproceedings{snpsd,
  title={Stroke-based neural painting and stylization with dynamically predicted painting region},
  author={Hu, Teng and Yi, Ran and Zhu, Haokun and Liu, Liang and Peng, Jinlong and Wang, Yabiao and Wang, Chengjie and Ma, Lizhuang},
  booktitle={Proceedings of the 31st ACM International Conference on Multimedia},
  pages={7470--7480},
  year={2023}
}

@inproceedings{snp,
  title={Stylized neural painting},
  author={Zou, Zhengxia and Shi, Tianyang and Qiu, Shuang and Yuan, Yi and Shi, Zhenwei},
  booktitle={Proceedings of the IEEE/CVF Conference on Computer Vision and Pattern Recognition},
  pages={15689--15698},
  year={2021}
}

@article{DifferentiableSketching,
  title={Differentiable drawing and sketching},
  author={Mihai, Daniela and Hare, Jonathon},
  journal={arXiv preprint arXiv:2103.16194},
  year={2021}
}

@article{diffvg,
  title={Differentiable vector graphics rasterization for editing and learning},
  author={Li, Tzu-Mao and Luk{\'a}{\v{c}}, Michal and Gharbi, Micha{\"e}l and Ragan-Kelley, Jonathan},
  journal={ACM Transactions on Graphics (TOG)},
  volume={39},
  number={6},
  pages={1--15},
  year={2020},
  publisher={ACM New York, NY, USA}
}

@article{vectorized,
  title={Vectorized Region Based Brush Strokes for Artistic Rendering},
  author={Prudviraj, Jeripothula and Jamwal, Vikram},
  journal={arXiv preprint arXiv:2506.09969},
  year={2025}
}

@inproceedings{painterly,
  title={Painterly rendering with curved brush strokes of multiple sizes},
  author={Hertzmann, Aaron},
  booktitle={Proceedings of the 25th annual conference on Computer graphics and interactive techniques},
  pages={453--460},
  year={1998}
}

@article{gpu,
  title={GPU-friendly Stroke Expansion},
  author={Levien, Raph and Uguray, Arman},
  journal={Proceedings of the ACM on Computer Graphics and Interactive Techniques},
  volume={7},
  number={3},
  pages={1--29},
  year={2024},
  publisher={ACM New York, NY, USA}
}

@article{offsets,
  title={Offsets of two-dimensional profiles},
  author={Tiller, Wayne and Hanson, Eric G},
  journal={IEEE Computer Graphics and Applications},
  volume={4},
  number={9},
  pages={36--46},
  year={2007},
  publisher={IEEE}
}

@inproceedings{interactive-hb,
  title={Interactive GPU-based procedural heightfield brushes},
  author={De Carpentier, Giliam JP and Bidarra, Rafael},
  booktitle={Proceedings of the 4th International Conference on Foundations of Digital Games},
  pages={55--62},
  year={2009}
}

@inproceedings{impasto,
  title={IMPaSTo: A realistic, interactive model for paint},
  author={Baxter, William and Wendt, Jeremy and Lin, Ming C},
  booktitle={Proceedings of the 3rd International Symposium on Non-photorealistic Animation and Rendering},
  pages={45--148},
  year={2004}
}

@article{wetbrush,
  title={Wetbrush: GPU-based 3D painting simulation at the bristle level},
  author={Chen, Zhili and Kim, Byungmoon and Ito, Daichi and Wang, Huamin},
  journal={ACM Transactions on Graphics (TOG)},
  volume={34},
  number={6},
  pages={1--11},
  year={2015},
  publisher={ACM New York, NY, USA}
}

@article{yang2024depth,
  title={Depth anything v2},
  author={Yang, Lihe and Kang, Bingyi and Huang, Zilong and Zhao, Zhen and Xu, Xiaogang and Feng, Jiashi and Zhao, Hengshuang},
  journal={Advances in Neural Information Processing Systems},
  volume={37},
  pages={21875--21911},
  year={2024}
}

@article{3dgs,
  title={3D Gaussian splatting for real-time radiance field rendering.},
  author={Kerbl, Bernhard and Kopanas, Georgios and Leimk{\"u}hler, Thomas and Drettakis, George},
  journal={ACM Trans. Graph.},
  volume={42},
  number={4},
  pages={139--1},
  year={2023}
}

@InProceedings{div2k,
	author = {Agustsson, Eirikur and Timofte, Radu},
	title = {NTIRE 2017 Challenge on Single Image Super-Resolution: Dataset and Study},
	booktitle = {The IEEE Conference on Computer Vision and Pattern Recognition (CVPR) Workshops},
	month = {July},
	year = {2017}
}

@inproceedings{chen2023editable,
  title={Editable image geometric abstraction via neural primitive assembly},
  author={Chen, Ye and Ni, Bingbing and Chen, Xuanhong and Hu, Zhangli},
  booktitle={Proceedings of the IEEE/CVF International Conference on Computer Vision},
  pages={23514--23523},
  year={2023}
}

@inproceedings{chen2024towards,
  title={Towards high-fidelity artistic image vectorization via texture-encapsulated shape parameterization},
  author={Chen, Ye and Ni, Bingbing and Liu, Jinfan and Huang, Xiaoyang and Chen, Xuanhong},
  booktitle={Proceedings of the IEEE/CVF Conference on Computer Vision and Pattern Recognition},
  pages={15877--15886},
  year={2024}
}

@inproceedings{chen2025easy,
  title={Easy-editable image vectorization with multi-layer multi-scale distributed visual feature embedding},
  author={Chen, Ye and Hu, Zhangli and Zhao, Zhongyin and Zhu, Yupeng and Shi, Yue and Xiong, Yuxuan and Ni, Bingbing},
  booktitle={Proceedings of the Computer Vision and Pattern Recognition Conference},
  pages={23345--23354},
  year={2025}
}
}

\clearpage
\setcounter{page}{1}
\maketitlesupplementary

\section{Physically-Grounded Height Field Construction per Stroke}
To achieve physically plausible rendering, our method incorporates a height field reconstruction loss (Sec.~\ref{sec:renderer}). We precompute a reference height field \( H_{gt} \) that encodes both global depth relations and local surface undulations. During rendering, we splat paint strokes to generate the color field \( I_{render} \) and, concurrently, form a height field \( H_{render} \) from the per-stroke height parameters \( h_i \) at negligible computational overhead. We then minimize the MSE loss between \( H_{render} \) and \( H_{gt} \), formulated as:
\begin{equation}\small
\mathcal{L}_{height} = \| H_{\text{render}} - H_{\text{gt}} \|_2,
\end{equation}
which guides each \(h_i\) toward a physically interpretable surface elevation.

\begin{figure}[htbp]
    \centering
    \includegraphics[width=1.0\linewidth]{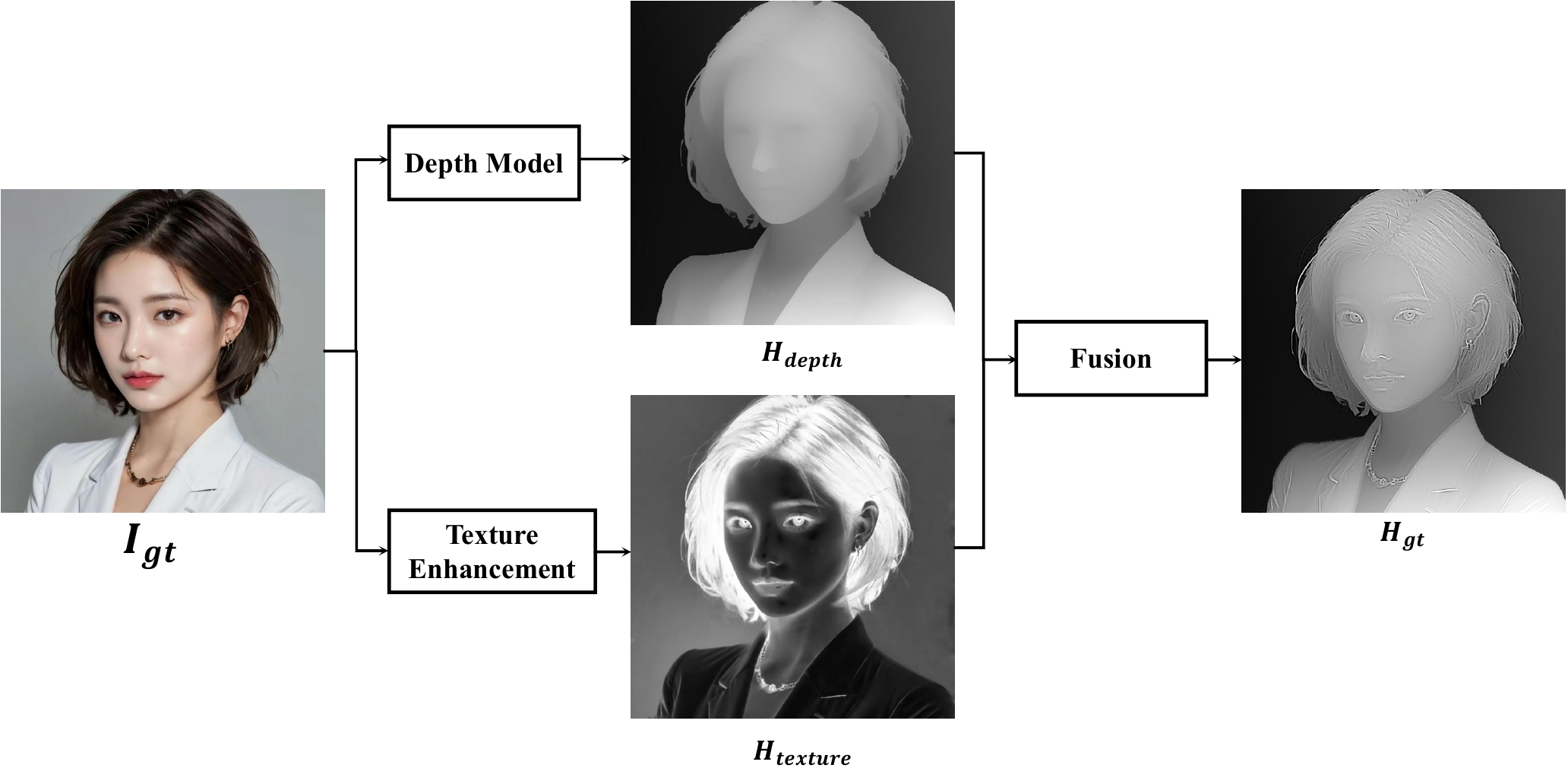}
    \caption{Overview of our height field construction pipeline. We fuse a depth-based height field, representing global scene structure, with a texture-based height field, capturing local surface details, to generate the final physically-plausible height field \(H_{\text{gt}}\).}
    \label{fig:height_framework}
\end{figure}

The target height field \(H_{\text{gt}}\) is constructed following a structure–texture decomposition principle, where depth supplies coarse ordering of objects in the scene and the image texture contributes high-frequency surface relief (see Fig.~\ref{fig:height_framework}). We first obtain a \textbf{depth-based height field} \(H_{\text{depth}}\) from a monocular depth estimation model (e.g., Depth Anything V2). In parallel, we derive a \textbf{texture-based height field} \(H_{\text{texture}}\) by converting the input image \(I_{\text{gt}}\) to the CIELAB color space and extracting the high-frequency component from its luminance channel. We fuse these two complementary components into the final \(H_{\text{gt}}\):
\begin{equation}\small
H_{\text{gt}} = \lambda_h H_{\text{depth}} + (1-\lambda_h) H_{\text{texture}},
\end{equation}
where $\lambda_h$ is a blending weight, which is typically set to 0.6.

\section{Lighting for Painterly Rendering}
\begin{figure*}[htbp]
    \centering
    \includegraphics[width=0.85\linewidth]{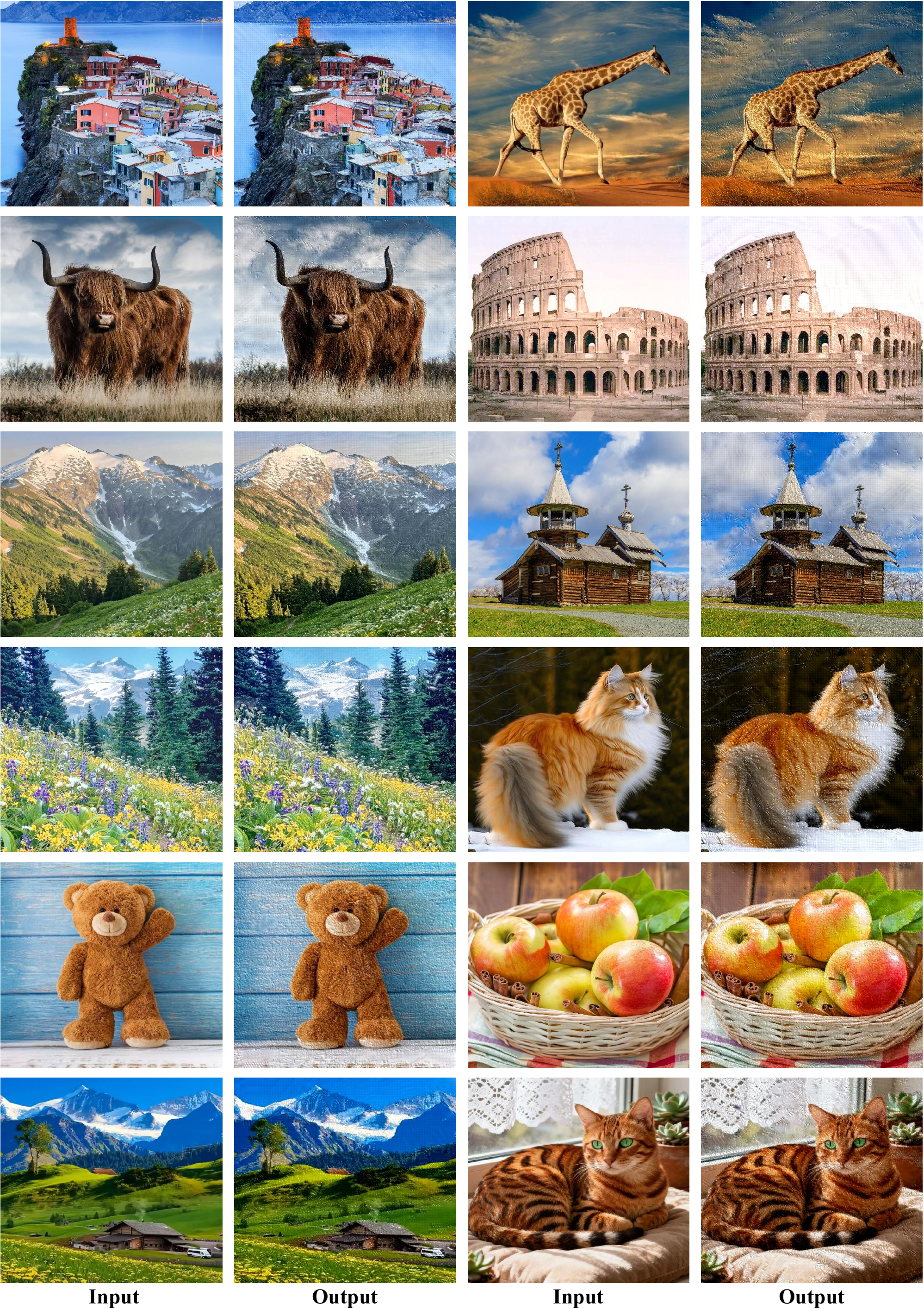}
    \caption{Relighting results using our height field representation. Our rendered result with canvas texture and impasto effects under directional lighting. The height field captures both macroscopic paint accumulation and microscopic canvas weave.}
    \label{fig:lighting_results}
\end{figure*}

To reproduce the tactile character of real oil paintings, we construct a height-field representation that models both the canvas substrate and the impasto accumulation of paint (see Fig.~\ref{fig:lighting_results}). For each stroke, we define a per-pixel height contribution \(h_i(\mathbf{x})\), which is procedurally perturbed to emulate bristle-induced ridges and grooves. These stroke-level height fields are composited with the canvas micro-geometry using the same front-to-back alpha blending mechanism described in Sec.~\ref{sec:renderer}, ensuring that height accumulation follows the identical ordering and transparency behavior as color splatting.

The canvas micro-geometry is synthesized procedurally using composite noise. We construct a base albedo field
\begin{align}
C(\mathbf{x}) 
&= C_0 + \Delta C \Big[
    \operatorname{fbm}(\mathbf{x}) \nonumber \\
&\quad\;\; + w_{\text{weave}}
    \sin(20\,\mathbf{x}\!\cdot\!d_x)\,
    \sin(20\,\mathbf{x}\!\cdot\!d_y)
\Big],
\end{align}
where \(C_0=(0.8,0.75,0.7)\), \(\Delta C=0.3\), and \(d_x\!\perp\! d_y\) specify the weave directions. Here \(\operatorname{fbm}(\cdot)\) denotes fractional Brownian motion noise, which introduces multi-scale stochastic variation typical in procedural texture synthesis. We find that \(w_{\text{weave}}=0.45\) produces a visually plausible balance between FBM noise and weave detail across all canvas resolutions.
The canvas height \(h_c(\mathbf{x})\) is derived from the luminance of \(C(\mathbf{x})\), producing fabric-like micro-relief.

Each stroke’s height is enhanced through a procedural impasto model. The perturbed height map is defined as
\begin{equation}\small
\tilde{h}_i(\mathbf{x}) 
= h_i(\mathbf{x}) 
+ A(r_i) \sum_{k=1}^{2} w_k 
  \sin\!\Big(\kappa_k f_0 r_i^{-1} \, l_i(\mathbf{x}) + \phi_k\Big),
\end{equation}
where \(r_i\) is the stroke radius, \(l_i(\mathbf{x})\) the coordinate along the stroke direction, 
\(A(r_i)=0.1 r_i\) controls amplitude, \(f_0=0.5\) is a base frequency, 
\((w_1,w_2)=(0.65,0.35)\) are harmonic weights, 
\((\kappa_1,\kappa_2)=(1.0,1.8)\) specify frequency ratios, 
and \(\phi_k\) are random per-stroke phases to avoid repetition. 
This formulation generates layered ridge patterns characteristic of real brushwork.

To couple paint and canvas geometry, we apply thickness-aware modulation:
\begin{align}
\hat{h}_i(\mathbf{x})
&= \tilde{h}_i(\mathbf{x})\,
\Big[ 1 + \big(1 - \gamma(\tilde{h}_i(\mathbf{x}))\big)\, h_c(\mathbf{x}) \Big], \\
\gamma(t) &= \alpha \min(t/h_t, 1),
\end{align}
using \(\alpha = 0.8\) and \(h_t = 40\). 
Thin strokes inherit more canvas-induced roughness through \(h_c(\mathbf{x})\), while thick impasto strokes (\(\gamma \to 1\)) gradually override the canvas texture.

The final height field \(H(\mathbf{x})\) is obtained by splatting the modulated stroke heights \(\hat{h}_i(\mathbf{x})\) using the same front-to-back compositing as color (Eq.~\ref{accu_color}), with height values replacing the per-segment color.  
This shared compositing mechanism ensures consistent ordering and enables height accumulation without additional computational cost.

Surface normals are computed from the resulting height field via spatial gradients:
\begin{equation}\small
\mathbf{n}(\mathbf{x}) = \operatorname{normalize}\big([-s\,\nabla H(\mathbf{x}), 1]^\top\big),
\end{equation}
where \(s\) is a height-to-slope scaling factor converting pixel-level height variation into shading-consistent surface slopes.

Shading is performed using a GGX microfacet reflectance model:
\begin{equation}\small
L = \frac{\rho}{\pi}L_d +
\frac{D\,F\,G}{4(\mathbf{n}\!\cdot\!\mathbf{v})
               (\mathbf{n}\!\cdot\!\mathbf{l})} L_s,
\end{equation}
with roughness \(\alpha_r=0.3\), Fresnel base reflectance \(F_0=0.08\) (typical for non-metallic pigments), and lighting intensities \((L_d, L_s)=(1.0,0.8)\). 
This reflectance model accentuates both canvas microstructure and impasto depth under directional illumination.

\section{Theoretical Analysis of the Bézier Intermediate Representation}
\label{sec:supp_bezier_analysis}

In our stroke-based rendering (SBR) framework, we parameterize strokes using a set of vertices $\theta$. Optimizing $\theta$ directly in the pixel domain is ill-posed and prone to high-frequency noise. To address this, we employ a Bézier curve as a differentiable geometric proxy. This process can be formally understood as a variant of \textit{Projected Gradient Descent}, where the optimization trajectory is constrained to the manifold of smooth cubic polynomials. 

Without loss of generality, we present the analysis for a single cubic Bézier segment. In practice, complex strokes are modeled as piecewise chains of such segments (splines), where this analysis applies locally to each segment.

\subsection{The Bézier Proxy as a Projection Operator}

Let a stroke segment be represented by $N$ vertices $P \in \mathbb{R}^{N \times d}$, flattened into a parameter vector $\theta = \operatorname{vec}(P) \in \mathbb{R}^{Nd}$. We define a differentiable rendering loss $L(\theta)$. 

The projection begins by fitting a cubic Bézier curve to the vertices. We assign scalar parameters $\{t_i\}_{i=0}^{N-1}$ based on cumulative chord length and construct the Bernstein basis matrix $B \in \mathbb{R}^{N \times 4}$. The control points $C \in \mathbb{R}^{4 \times d}$ are obtained via least-squares fitting:
\begin{equation}\small
C = (B^\top B)^{-1} B^\top P.
\end{equation}
To express this rigorously in vector form, we utilize the Kronecker product $\otimes$. Let $c = \operatorname{vec}(C) \in \mathbb{R}^{4d}$. The fitting operation is a linear projection:
\begin{equation}
    c = F \theta, \quad \text{where} \quad F = \left( (B^\top B)^{-1} B^\top \right) \otimes I_d.
\end{equation}
Here, $I_d$ is the $d \times d$ identity matrix, ensuring the projection applies independently to each spatial dimension.

Following projection, we define a reconstruction operator $S$ that samples the curve at fixed intervals to produce a refined polyline $\tilde{\theta}$. The composite operator $M = S F$ projects the arbitrary vertex vector $\theta$ onto the Bézier subspace. The optimization objective becomes $\tilde{L}(\theta) = L(M\theta)$.

\subsection{Decoupling Normal and Tangential Flow}

A critical aspect of our formulation is the handling of the parameterization $\{t_i\}$. Since $t_i$ depends on the vertex positions $P$, the matrix $M$ is technically a function of $\theta$. By the chain rule, the exact gradient is:
\begin{equation}\small
\nabla_\theta \tilde{L} = M^\top \nabla_{\tilde{\theta}} L + (\nabla_\theta M)^\top \nabla_{\tilde{\theta}} L.
\end{equation}
However, we explicitly discard the second term $(\nabla_\theta M)^\top$. This is not merely a computational simplification, but a deliberate design choice to \textbf{decouple geometric deformation from parameterization drift}.
In geometric optimization, the gradient of the loss typically contains two components: a \textit{normal} component that changes the shape, and a \textit{tangential} component that merely slides vertices along the curve without altering geometry. The term $(\nabla_\theta M)^\top$ couples these effects. By treating the parameterization $\{t_i\}$ (and thus $M$) as locally frozen, we enforce a robust inductive bias: the optimizer focuses solely on minimizing the rendering loss via shape deformation, preventing pathological vertex clustering or tangential drift.

\subsection{Spectral Analysis: Subspace Constraint}

The smoothing effect of our method stems from the rank-deficient nature of the projection $F$. The matrix $B$ maps the 4-dimensional control point space to the $N$-dimensional vertex space.
Consider the Singular Value Decomposition (SVD) of the basis matrix $B$. Since $B$ has at most rank 4, the projection operator $F$ has a null space of dimension $N-4$ (assuming $N > 4$). 

This implies that our proxy acts as a strict \textbf{low-rank geometric filter}. Any high-frequency variation in the vertex gradient $\nabla_\theta L$ that falls into this null space is mathematically eliminated during the backpropagation step $\nabla_\theta \tilde{L} \approx F^\top (\dots)$. Unlike a standard low-pass filter which attenuates frequencies based on a cutoff, the Bézier projection fundamentally restricts the degrees of freedom of the solution to the cubic polynomial subspace. This improves the condition number of the optimization problem by removing high-frequency noise directions from the loss landscape.

\subsection{Implementation Details and Robustness}
Our fitting procedure adapts to the stroke complexity to ensure numerical stability. For strokes with sparse vertices ($N \le \text{degree}+1$), global least-squares fitting is ill-conditioned. In these cases, we construct a Catmull--Rom spline through the vertices and convert each segment into an equivalent cubic Bézier curve to ensure $C^1$ continuity. For strokes with sufficient vertices ($N > \text{degree}+1$), we perform a global least-squares fit of an $n$-th degree Bézier curve. We solve the linear system $Bc \approx P$ using a normalized chord-length parameterization. Throughout optimization, we monitor the symmetric Chamfer distance between the polyline and the Bézier proxy to ensure the representation remains faithful.

\begin{figure}[htbp]
  \centering
  \includegraphics[width=1.00\linewidth]{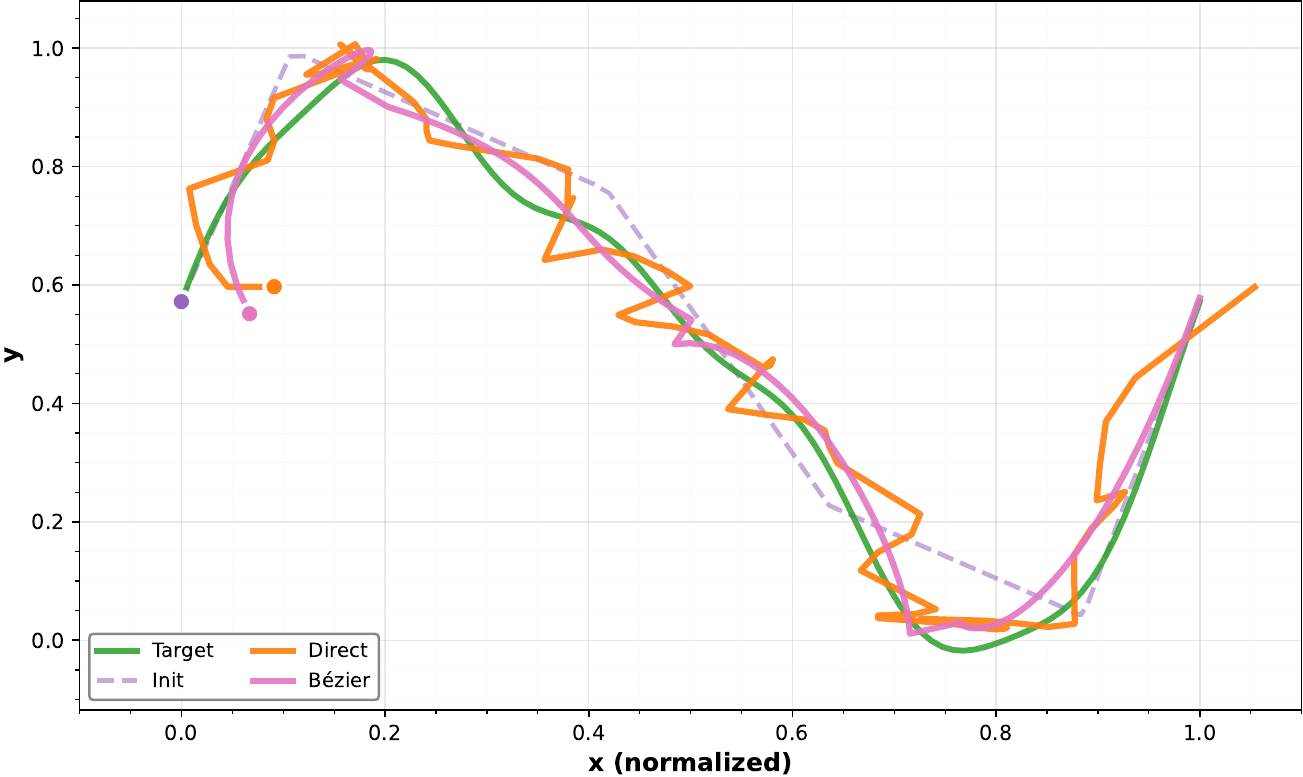} 
  \caption{\textbf{Trajectory comparison.} Direct optimization (orange) overfits to noise. Bézier proxy (magenta) projects the solution onto the cubic polynomial subspace.}
  \label{fig:supp-noise-bezier-traj}
\end{figure}

\begin{figure}[htbp]
  \centering
  \includegraphics[width=1.00\linewidth]{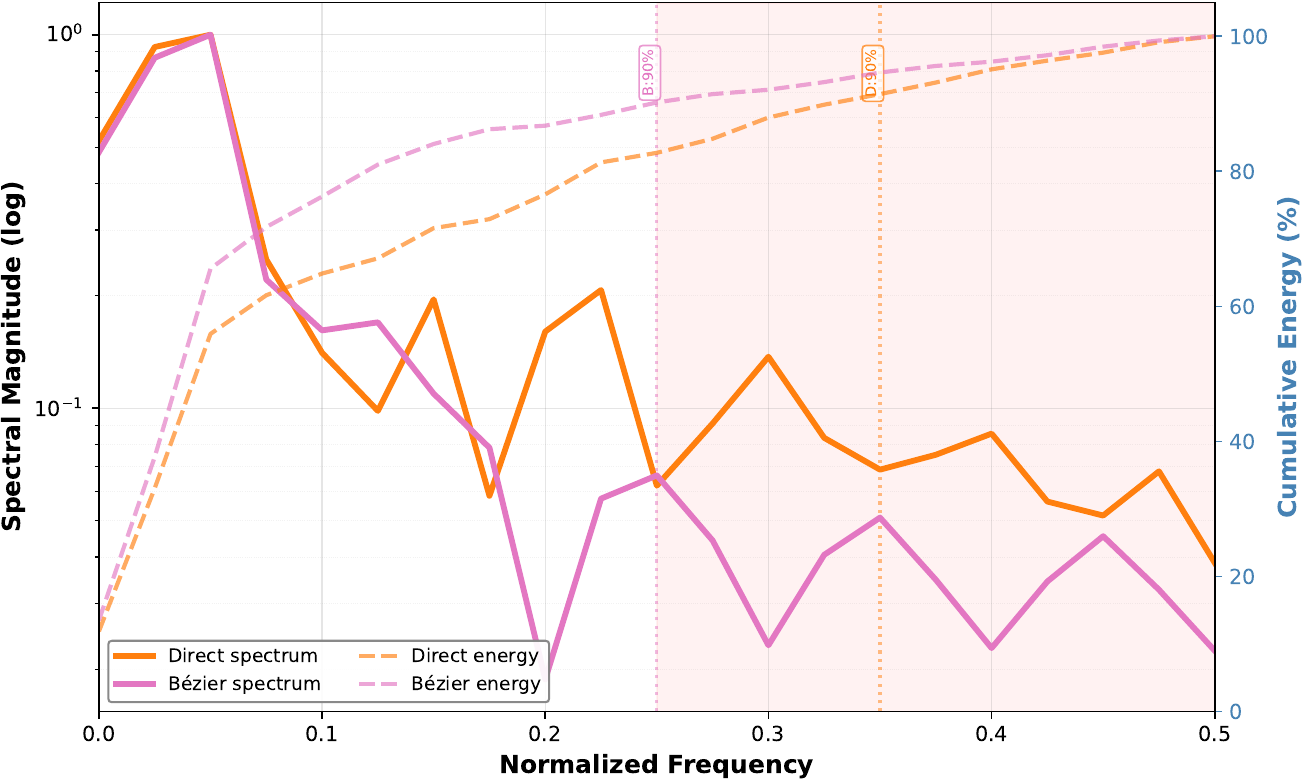} 
  \caption{\textbf{Gradient Spectrum.} Spectral analysis confirms that the proxy operator $M^\top$ nullifies gradient components associated with high-frequency spatial variation.}
  \label{fig:supp-noise-bezier-spec}
\end{figure}

To empirically demonstrate the smoothing and regularizing effect of the Bézier proxy, we conduct a toy experiment. We define a smooth ground-truth stroke and add i.i.d. Gaussian noise ($\mathcal{N}(0, 0.05^2)$) to its vertices, creating a noisy target. We then initialize a stroke and optimize its control points for 200 iterations (learning rate 0.3) to fit this noisy target, comparing two methods: (i) direct optimization of the vertices, and (ii) optimization through our Bézier proxy.

As shown in Fig.~\ref{fig:supp-noise-bezier-traj}, direct optimization quickly overfits to the noise, producing a visually jagged stroke. The Bézier proxy, however, ignores the high-frequency jitter and recovers a smooth curve that faithfully captures the underlying structure of the clean target. The gradient spectrum analysis (Fig.~\ref{fig:supp-noise-bezier-spec}) confirms our theoretical intuition: the Bézier proxy significantly dampens high-frequency components of the gradient update, thus stabilizing the optimization and promoting structurally sound results.

\end{document}